\documentclass[sigconf]{acmart}

\usepackage{booktabs} 
\usepackage[utf8]{inputenc}

\usepackage{hyphenat}
\usepackage{url}
\usepackage{tabularx}

\begin{document}
\title{Improving Small Object Proposals for Company Logo Detection}

\author{Christian Eggert}
\email{christian.eggert@informatik.uni-augsburg.de}

\author{Dan Zecha}
\email{dan.zecha@informatik.uni-augsburg.de}

\author{Stephan Brehm}
\email{stefan.brehm@informatik.uni-augsburg.de}

\author{Rainer Lienhart}
\email{rainer.lienhart@informatik.uni-augsburg.de}

\affiliation{%
  \institution{University of Augsburg}
  \streetaddress{Universitätsstr. 6a}
  \city{Augsburg} 
  \postcode{86199}
}

\renewcommand{\shortauthors}{C. Eggert et al.}

\begin{abstract}
Many modern approaches for object detection are two-staged pipelines.
The first stage identifies regions of interest which are then classified in the second stage.
Faster R-CNN is such an approach for object detection which combines both stages into a single pipeline.
In this paper we apply Faster R-CNN to the task of company logo detection.
Motivated by its weak performance on small object instances, we examine in detail both the proposal and the classification stage with respect to a wide range of object sizes.
We investigate the influence of feature map resolution on the performance of those stages.

Based on theoretical considerations, we introduce an improved scheme for generating anchor proposals and propose a modification to Faster R-CNN which leverages higher-resolution feature maps for small objects.
We evaluate our approach on the FlickrLogos dataset improving the RPN performance from $0.52$ to $0.71$ (MABO) and the detection performance from $0.52$ to $0.67$ (mAP).
\end{abstract}

%
%


\copyrightyear{2017}
\acmYear{2017}
\setcopyright{acmlicensed}
\acmConference{ICMR '17}{June 06-09, 2017}{Bucharest, Romania}
\acmPrice{15.00}
\acmDOI{10.1145/3078971.3078990}
\acmISBN{978-1-4503-4701-3/17/06}



\maketitle

\section{Introduction}
\label{sec:introduction}

Current object detection pipelines like Fast(er) R-CNN~\cite{girshick_2015_fastrcnn}~\cite{ren_2015_faster_rcnn} are built on deep neural networks whose convolutional layers extract increasingly abstract feature representations by applying previously learned convolutions followed by a non-linear activation function to the image.
During this process, the intermediate feature maps are usually downsampled multiple times using max-pooling.

This downsampling has multiple advantages:
\begin{enumerate}
\item It reduces the computational complexity of applying the model
\item It helps to achieve a certain degree of translational invariance of the feature representation
\item It also increases the receptive field of neurons in the deeper layers.
\end{enumerate}
The flipside of these advantages is a feature map which has a significantly lower resolution than the original image. 
As a result of this reduced resolution it is difficult to associate features with a precise location in the original image.

Despite this potential drawback, this approach has been extremely successful in the areas of image classification and object detection.
For most applications, pixel-accurate localization is not important.

In this paper we examine the suitability of feature representations from different levels of the feature hierarchy for the problem of company logo detection. 
Company logo detection is an application of object detection which attracts lots of commercial interest.
On a superficial level, company logo detection is nothing but a special case of general object detection.
However, company logos are rarely the objects which were intended to be captured when the picture was taken.
Instead, they usually happen to get caught in the picture by accident.
As a result, company logos tend to occupy a rather small image area.

Intersection over union (IoU) is the usual criterion by which the quality of the localization is assessed. 
By this measure, a detection which is off by a given amount of pixels has a greater influence on small object instances than large ones.
Therefore, small object instances require a more precise localization than large instances in order to be classified as correct detections.

Another problem is, that small objects typically are detected with a lower confidence score than large objects.
Experiments done by~\cite{eggert_saliency_guided_magnification_2016} show that this effect is not necessarily rooted in the low resolution of the objects and that detection performance can be improved by simply magnifying the test images.

Magnifying input images would also alleviate the former problem but this simple approach is not very appealing since the effort for applying the convolutions grows quadratically with the side length of the image.
This is especially true for company logo detection in which the object is typically small compared to the image, resulting in much unnecessary computation.

Our contributions are as follows:
\begin{enumerate}
\item We theoretically examine the problem of small objects at the proposal stage.
We derive a relationship which describes the minimum object size which can reasonably be proposed and provide a heuristic for choosing appropriate anchor scales.
\item We perform detailed experiments which capture the behavior of both the proposal and the classification stage as as a function of object size using features from different feature maps. 
Deeper layers are potentially able to deliver features of higher quality which means that individual activations are more specific to input stimuli than earlier layers. We show that in the case of small objects, features from earlier layers are able to deliver a performance which is on par with -- and can even exceed -- the performance of features from deeper layers.
\item We evaluate our observations on the well-known FlickrLogos dataset~\cite{Romberg_2011} in the form of an extension to the Faster R-CNN pipeline
\end{enumerate}
Since FlickrLogos has been originally been conceived as a benchmark for image retrieval we have re-annotated the dataset for the task of object detection\footnote{The updated annotations and evaluation script are made available here: \url{http://www.multimedia-computing.de/flickrlogos} }.

\section{Related Work}
\label{sec:related_work}
Object proposals traditionally have played an important role in DCNN-based object detection due to the high computational cost of applying a DCNN in a sliding window fashion.

R-CNN~\cite{girshick_2014_rcnn} was one of the first approaches which applied DCNNs to object detection.
External algorithms like Selective Search~\cite{uijlings_2013_selsearch} or Edge Boxes~\cite{zitnick_2014_edge_boxes} were used to generate regions of interest.
The DCNN would extract a separate feature representation of each ROI which was subsequently used for classification through a support vector machine~\cite{boser_1992_support_vector_machine}.

Fast R-CNN~\cite{girshick_2015_fastrcnn} is able to speed up object detection by not computing a separate feature representation for each ROI.
It applies the convolutional layers of a DCNN across the complete image, generating a single feature map.
For each ROI, an ROI-Pooling layer computes a fixed-dimensional representation from parts of the feature map, which is then used for classification.
Faster R-CNN~\cite{ren_2015_faster_rcnn} finally incorporates the generation of object proposals into the network itself by introducing a region proposal network (RPN) which operates on the same single feature map of the image.

Some approaches to object detection can manage without explicitly generating object proposals.
Two representatives of this class are YOLO~\cite{redmon_2016_yolo} and SSD~\cite{wei_2016_ssd}.
However, these algorithms are typically optimized for real-time object detection and usually do not perform as well on small object instances which limits their applicability for company logo detection.

Scale is a potentially bigger problem for Fast(er) R-CNN than for R-CNN since it does not rescale every ROI to a standard size.
Therefore, some efforts have been made to mitigate this problem:
\cite{eggert_saliency_guided_magnification_2016} use a multi-scale approach while avoiding to build a complete image pyramid by using the initial feature representation of small objects to decide which ROIs should be examined in more detail.
The corresponding image parts are packed into a new image which is magnified and a new feature map is being computed.

Other approaches build a multi-scale feature representation:
\cite{bell_inside_outside_net_2016} apply techniques like skip-pooling to create multi-scale feature representations.
They also consider encoding the context of an object using features obtained by an recurrent network.
Hypercolumns~\cite{hariharan_2016_hypercolumns} attempt to construct a single multi-scale feature representation by concatenating feature maps generated on different levels of the DCNN while \cite{shelhamer_2016_semantic_image_segmentation} combine predictions based on different feature maps to a single more accurate preduction.
Most similar to the network architecture used in our work are Feature Pyramid Networks~\cite{lin_2016_fpn} (FPNs) which compute hierarchical feature maps on different scales but introduce another path to the network which aggregates the different feature map into a multi-scale representation.

All of these approaches focus on multi-scale feature representations to increase the performance of the classifier.
However, in this work we focus on the limitations that are inherent in the choice of the anchor sets.
By this, we mean the impact of anchors on the RPN performance, assuming a perfect classifier.
Therefore the considerations in this paper are largely complementary to the previously mentioned approaches.

The specific problem of company logo detection with DCNNs has previously been studied by \cite{oliveira_2016_automatic_graphic_logo_detection, eggert_saliency_guided_magnification_2016} and \cite{bianco_2017_deep_learning_for_logo_recognition}.
All of these approaches use the above-mentioned R-CNN~\cite{girshick_2014_rcnn} or Fast R-CNN~\cite{girshick_2015_fastrcnn} pipelines using externally generated object proposals.
In this paper we apply RPNs to this task and study the problems that arise for small objects.

\section{Small Objects in Faster R-CNN}
\label{sec:small_objects_in_faster_r-cnn}

Current object detection pipelines usually consist of two stages:
First, they identify regions of interest (ROIs) from images.
These ROIs serve as an attention model and propose potential object locations which are more closely examined in the second stage.

For our experiments we use a re-implementation of the Faster R-CNN~\cite{ren_2015_faster_rcnn} approach.
Faster R-CNN extracts a feature representation of the image through a series of learned convolutions.
This feature map forms the basis of both the object proposal stage and the classification stage.
The first step is accomplished by a Region Proposal Network (RPN) which starts by generating a dense grid of anchor regions with specified size and aspect ratio over the input image.

For each anchor, the RPN -- which is a fully convolutional network -- predicts a score which is a measure of the probability of this anchor containing an object of interest.
Furthermore, the RPN predicts two offsets and scale factors for each anchor which are part of a bounding box regression mechanism which refines the object's location.
The refined anchors are sorted by score, subjected to a non-maximum suppression and the best scoring anchors are kept as object proposals which are fed into the second stage of the network.

At training time, anchors are divided into positive and negative examples, depending on the overlap they have with a groundtruth instance.
Typically, an anchor is considered to be a positive example if it has an IoU greater than $0.5$ with a groundtruth object.

\begin{figure}
\centering
\includegraphics[scale=1.0]{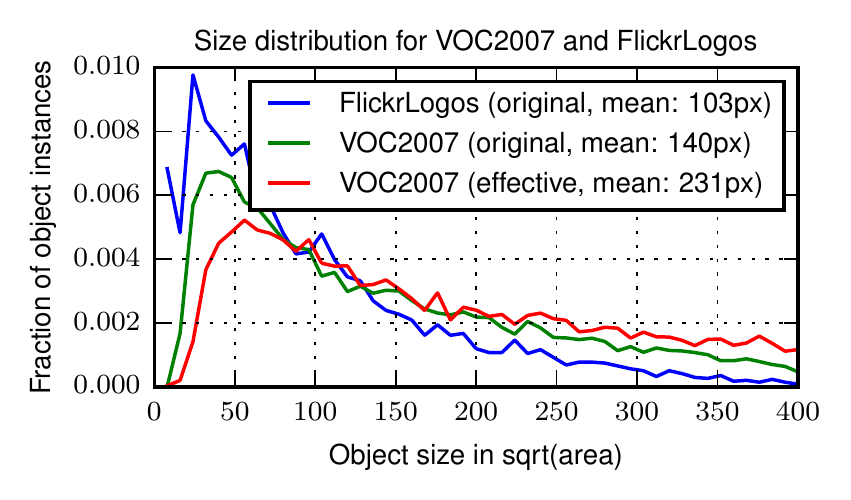}
\caption{Distribution of object instance sizes in VOC2007 and FlickrLogos. The effective size distribution -- induced by dynamic rescaling in Faster R-CNN -- is shifted towards larger object instances. }
\label{fig:flickr_sizedist_orig}
\end{figure}

Ren et. al~\cite{ren_2015_faster_rcnn} use anchors whose side length are powers of two, starting with $128$ pixels.
This choice of anchors delivers good results on datasets such as VOC2007~\cite{everingham_2010_voc} where the objects are typically relatively large and fill a sizeable proportion of the total image area. 
Furthermore, \cite{ren_2015_faster_rcnn} also dynamically re-scale input images to enlarge the objects:
Images are rescaled in a way so that their minimum side length is at least 600px but their maximum side length does not exceed 1000px.
This rescaling results in a size distribution which is distinct from the one of the original VOC2007 dataset.
Both the original and the resulting size distribution are illustrated in Figure~\ref{fig:flickr_sizedist_orig}.

Upscaling the input images typically improves the detection performance which is interesting, since upscaling cannot introduce new information.
This has already been noted by \cite{eggert_saliency_guided_magnification_2016} who attribute this property to the receptive field of the network.

Figure~\ref{fig:flickr_sizedist_orig} also shows the size distribution of the FlickrLogos~\cite{Romberg_2011} dataset.
The average object size is quite small compared with the average side length of the images (which is typically around 1000 pixels).
Rescaling the images so that logo instances are on a comparable scale to the VOC2007 dataset would result in huge images.
This means that upscaling of input images is typically not feasible for company logo detection.

Figure~\ref{fig:flickr_sizedist_orig} also makes it clear, that an anchor of side length of $128px$ -- which is the smallest anchor scale used in the default Faster R-CNN approach -- is inadequate to cover the range of object sizes.
In order to counter this problem one could simply add additional anchors using the same powers-of-two scheme used by~\cite{ren_2015_faster_rcnn}.
However, we show that this scheme leads to difficulties -- particularly for small objects -- as it might fail to generate an anchor box with sufficient overlap.

\begin{figure}
\centering
\includegraphics[scale=1.0]{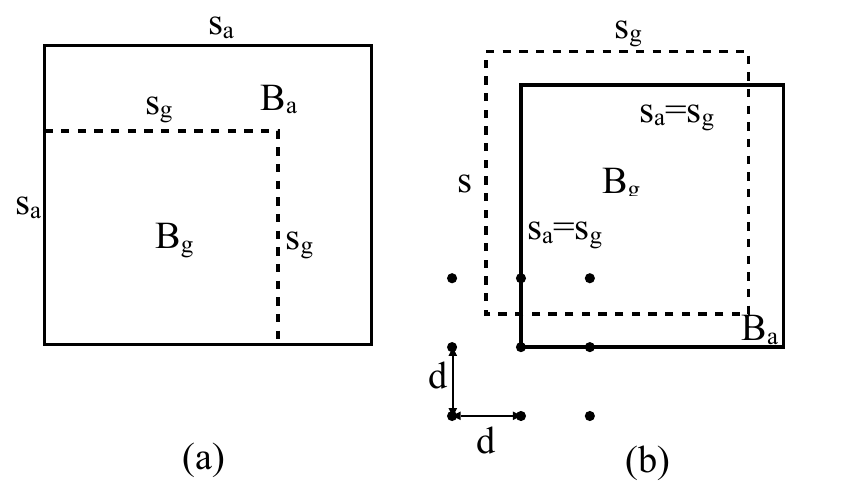}
\caption{(a) IoU can be expressed as the ratio of bounding box areas in the case of aligned bounding boxes of equal aspect ratio. (b) Worst case displacement of two bounding boxes of equal size when anchors are sampled with stride $d$}
\label{fig:bounding_box_vs_anchor_size}
\end{figure}

To illustrate the problem we will imagine an object proposal system which has learned to perfectly predict the concept of a ROI as taught during the training phase.
This means, we imagine a classifier which for every possible anchor is able to predict perfectly, whether the anchor has an IoU with a groundtruth instance of at least $0.5$.

Assuming such a classifier, consider the situation in Figure~\ref{fig:bounding_box_vs_anchor_size}a:
We assume a quadratic groundtruth bounding box $B_{g}$ of side length $s_{g}$ and a quadratic anchor box $B_{a}$ of side length $s_{a}$.
Furthermore we will assume w.l.o.g. that $s_{g} \leq s_{a}$ and that both side lengths are related through a scaling factor $\alpha \geq 1$ by $s_{a} = \alpha s_{g}$.
Under these conditions we can move $B_{g}$ anywhere inside of $B_{a}$ without changing the IoU.

In this case we can express the IoU as the ratio between the areas enclosed by these boxes:
\begin{equation}
t \leq IoU(B_{g},B_{a}) = \frac{|B_{g} \cap B_{a}| }{ |B_{g} \cup B_{a}| } = \frac{s_{g}^2}{s_{a}^2} = \frac{1}{\alpha^2}
\end{equation}

In order for an anchor box to be classified as positive example we require the IoU to exceed a certain threshold $t$. 
It follows that for $\alpha > \sqrt{t}^{-1}$ an anchor is unable to cover a groundtruth box with sufficient overlap to be classified as a positive example.
The same relationship holds for non-quadratic anchors -- provided the aspect ratio of groundtruth boxes and anchor boxes match.

Therefore, the side length of anchor boxes of neighboring scales $s_{a_1}$ and $s_{a_2}$ should be related by $s_{a_2} = \sqrt{t}^{-1} s_{a_1}$.

For the previous considerations we assume that there exists an anchor position at which the corner of an anchor is completely aligned with the groundtruth instance.
In practice this is not true since the feature map of the network upon which the RPN is based usually has a much smaller resolution than the original image.
A downsampling factor $d^{-1}$ between the original image and the feature map effectively results in a grid of anchors with stride $d$.

To examine the influence of feature map resolution on the RPNs potential to identify small object instances, we consider the situation in Figure~\ref{fig:bounding_box_vs_anchor_size}b.
We assume a quadratic groundtruth instance $B_g$ and the existence of an anchor box $B_a$ of identical scale and aspect ratio.
In the worst case, both boxes are displaced against each other by a distance of $\frac{d}{2}$.
The IoU between these boxes can be expressed by:
\begin{equation}
IoU(B_{g},B_{a}) = \frac{(s_g - \frac{d}{2})^2}{(s_g - \frac{d}{2})^2 + 2(2\frac{d}{2}(s_g-\frac{d}{2}) + \frac{d^2}{4})}
\end{equation}

Solving $t \leq IoU(B_{g},B_{a})$ for $s_g$ while assuming $d > 0$ and $0 < t < 1$ and ignoring negative solutions for this quadratic expression, we obtain the following relationship for the minimum detectable object size:
\begin{equation}
\frac{d(t+1) + d\sqrt{2t(t+1)}}{2-2t} \leq s_g
\end{equation}

For the VGG16~\cite{simonyan_2015_very_deep_networks} architecture, which is commonly used as basis for Faster R-CNN, $d=16$.
Assuming $t=0.5$, this translates into a minimum detectable object size of $s_g \approx 44px$.
This suggests that for the small end of our size distribution a feature map of higher resolution is needed.
For the \emph{conv4} feature map ($d=8$) the minimum detectable object size is given by $s_g \approx 22px$.
Since we do not expect to reliably classify objects smaller than 30px we use the next power of two as smallest anchor size.

Making use of our previous result we choose as our anchor set $\mathcal{A} = \left\lbrace 32, 45, 64, 90, 128, 181, 256 \right\rbrace$ since we follow the recommendation of \cite{ren_2015_faster_rcnn} and set $t=0.5$.

\subsection{Region Proposals of small objects}
\label{sec:region_proposals_with_small_objects}
We want to evaluate the effectiveness of RPNs for different object sizes.
The primary measure of an RPN's quality is the mean average best overlap (MABO). 
It measures the RPN's ability to generate at least one proposal region for each object with high overlap.
If $\mathcal{C}$ represents the set of object classes, $G_c$ the set of groundtruth objects of a particular class $c \in \mathcal{C}$ and $\mathcal{L}$ the set of object proposals, we can evaluate the performance of the RPN for a particular class $c$ via its average best overlap $ABO(c)$ given by:
\begin{equation}
ABO(c) = \frac{1}{|G_c|} \sum \limits_{g \in G_c} \max\limits_{l \in L} IoU(g,l)
\end{equation}

where $IoU(g,l)$ is the intersection over union between the groundtruth item $g$ and the proposal $l$. 
The MABO is the mean over all ABO values for each object class.

In order to examine the influence of object size on the performance of the RPN, we create differently scaled synthetic variants of the FlickrLogos~\cite{Romberg_2011} dataset by applying the following algorithm to each image:

We start by selecting the point which has the maximum distance between two non-overlapping groundtruth bounding boxes.
This point defines two axes along which the image is to be partitioned into four parts.
We ensure that the axes of the split do not intersect with any other groundtruth items.
If no such split can be found, the image is discarded.
For each of the resulting partitions which contain more than one groundtruth item, the process is applied recursively.
After applying this algorithm, each image contains only a single object instance which is then rescaled to match the desired target size.

Using this algorithm we create 11 differently scaled versions of the test set which we call $F_{test,x}$ where $x \in \left\lbrace 10*i+20 | i = 0 \hdots 10 \right\rbrace$ represents the target object size, measured as square root of the object area.
Additionally, we create a single training dataset $F_{train}$ in which the objects are scaled in such a way that the square root of the object area is distributed evenly in the interval $[20\text{px}, 120\text{px}]$.

\begin{figure}
\centering
\includegraphics[scale=1.0]{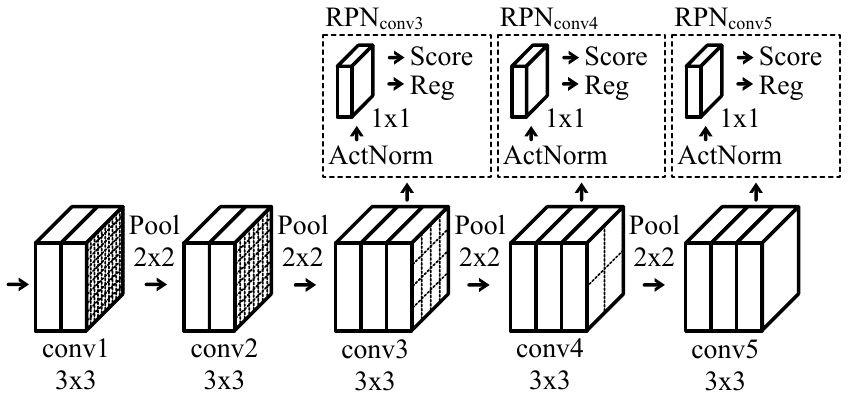}
\caption{Template for the construction of our RPNs. The RPN uses the same number of channels as the preceeding feature map and outputs object proposals as a grid of predictions with the same resolution.}
\label{fig:rpn_architecture_template}
\end{figure}

In order to observe the performance of the RPN for different layers we create three RPNs: $RPN_{conv3}$, $RPN_{conv4}$ and $RPN_{conv5}$ based on the VGG16~\cite{simonyan_2015_very_deep_networks} architecture used by \cite{ren_2015_faster_rcnn} and attach RPN modules to the \emph{conv3}\footnote{\emph{conv3} refers to the output of the last layer of the \emph{conv3} block which is \emph{conv3\_3} when using the naming convention of \cite{simonyan_2015_very_deep_networks}}, \emph{conv4} and \emph{conv5} layers, respectively.
The template for this network is illustrated in Figure~\ref{fig:rpn_architecture_template}.
Each network is trained separately with only one of the RPN modules active at a time.

The features are passed through a normalization layer which normalizes the activations to have zero-mean and unit-variance.
This is similar to batch normalization~\cite{ioffe_2015_batch_normalization}. 
However, we normalize the activations with respect to the training set and not with respect to the current batch as in \cite{ioffe_2015_batch_normalization}.
We do this so that we can easily use an off-the-shelf Imagenet~\cite{deng_imagenet_2009}-pretrained VGG16 network.
Those pre-trained models usually have the property that the variance of activations decreases from layer to layer as the data progresses through the network.
This property makes it hard to make certain changes to the network architecture. 
For example, adding additional branches of different depths will result in differently scaled activations in each branch which in turn leads to different effective learning rates in each branch.
This normalization scheme circumvents this problem.

We place a standard RPN on top of this feature normalization which consists of a $3 \times 3$ convolution using the same number of channels than the preceeding layer.
The output of this RPN is then used in two additional convolutional layers which predict anchor scores and regressors (see \cite{ren_2015_faster_rcnn} for details).
In the case of $RPN_{conv3}$ we use the features from the \emph{conv3} layer for predicting bounding boxes.

\begin{figure*}
\centering
\includegraphics[scale=1.0]{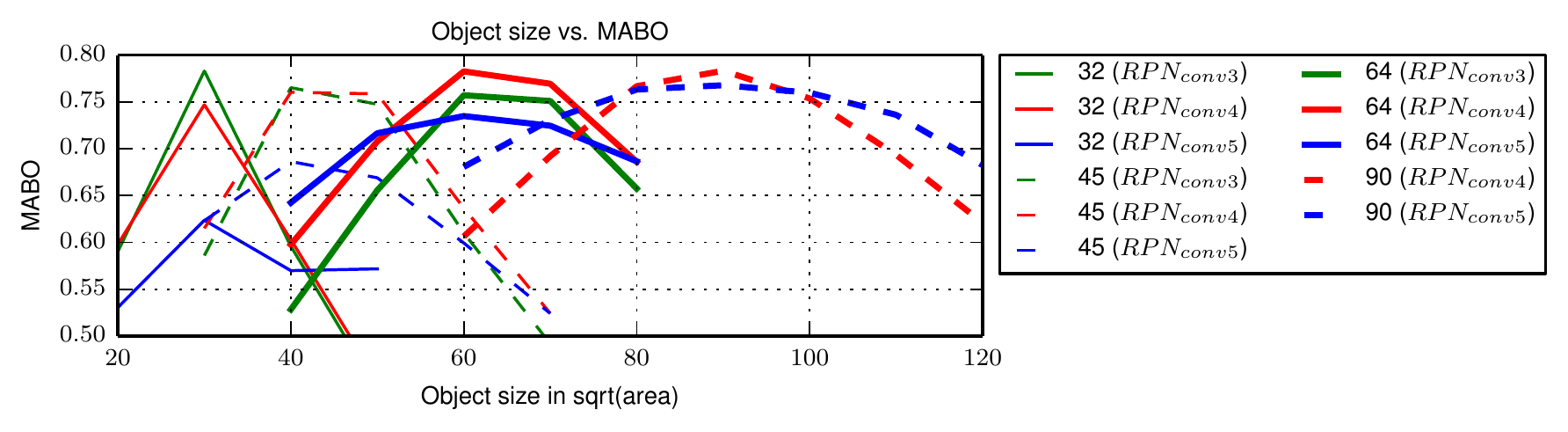}
\caption{RPN performance (MABO) for different anchor sizes as a function of object size. (green) performance of \emph{conv3}, (red) performance of \emph{conv4}, (blue) performance of \emph{conv5}. Proposals for small objects can be generated more effectively by earlier layers while the performance of the \emph{conv5} layer drops noticeably. }
\label{fig:result_rpn_performance_by_anchor_and_size}
\end{figure*}

We fine-tune each of our RPNs on the $F_{train}$ dataset for 40000 iterations with an initial learning rate of $\mu = 0.001$ on our set of anchors $\mathcal{A}$. 
The learning rate is decreased by a factor of $\gamma = 0.1$ after 30000 iterations.
We then evaluate the trained RPNs on the different $F_{test,x}$ datasets while only considering the outputs for a single anchor at a time.
As a result we are able to plot how effective the different feature maps are at predicting object proposals of a given size.
Figure \ref{fig:result_rpn_performance_by_anchor_and_size} shows the result of this experiment.
Each point on the abscissa represents the result of an experiment with the corresponding $F_{test,x}$ dataset while the ordinate reports the performance for this experiment as MABO.

Figure \ref{fig:result_rpn_performance_by_anchor_and_size} shows that for small objects the \emph{conv5} feature map delivers results which are noticably inferior than the results generated by the \emph{conv3} or \emph{conv4} feature maps.

Another observation to be made is that earlier feature maps deliver a more localized response for every anchor than the \emph{conv5} feature map. 
This manifests itself in a steeper performance drop as the object size moves away from the ideal anchor size.
This is a consistent pattern over all examined object sizes:
Even medium sized objects with a side length between 80px and 100px are better predicted by the \emph{conv4} feature map. 
However, this in only true if the object size closely matches the anchor size.
The \emph{conv5} feature map is able to deliver a more stable performance over a larger range of object sizes.


\subsection{ROI Classification of small objects}
\label{sec:roi_classification_of_small_objects}
After identifying ROIs, Faster R\-CNN predicts a score and bounding box regressants for each ROI and for every class.
In the original approach, this stage re-uses the previously computed \emph{conv5} feature map which was used to generate the object proposals.
An ROI-Pooling~\cite{girshick_2015_fastrcnn} layer projects the ROI coordinates identified by the RPN onto the feature map using the downsampling factor of the network.
The corresponding area of the feature map is converted into a fixed-dimensional representation with a pre-determined spatial resolution (usually $7 \times 7$).
Each of these feature representations is then fed into several fully connected layers for classification and class-specific bounding box regression.

\begin{figure}
\centering
\includegraphics[scale=1.0]{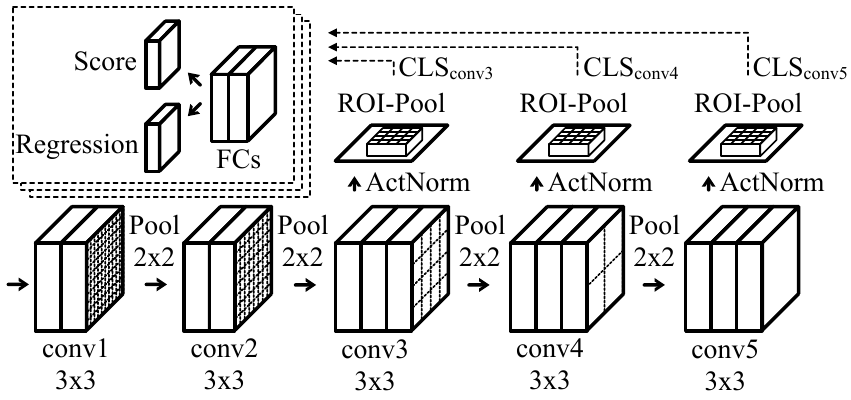}
\caption{Template for the construction of our classification networks. The fully connected layers of the classification pipeline are attached to an ROI-Pooling layer operating on a specific feature map.}
\label{fig:cls_architecture_template}
\end{figure}

We perform an analysis of the performance of the classification stage by object size which is similar to our analysis of the RPN.
Unlike RPNs, where each anchor by virtue of its size and the overlap criterion self-selects appropriate training examples, the classification stage does not have this property.
We therefore need to be careful about the size distribution in the training set.

For the scope of this paper we are interested in the maximum performance each feature map can deliver for a specific object size.
In order to avoid any effects from size distribution we ideally want a separate training set for each test set $F_{test,x}$.
To reduce the training effort, we combine multiple sizes into a single training set.
For this purpose we generate four training sets $F_{train,a,b}$ where $a$ represents the minimum object size and $b$ the maximum object size as the square root of the object area.
We choose 
\begin{equation*}
(a,b) \in \left\lbrace (20px,60px), (40px,80px), (60px,100px), (80px,120px) \right\rbrace
\end{equation*} 
to adequately cover the range of small objects in the FlickrLogos dataset (see Figure \ref{fig:flickr_sizedist_orig}).

Similar to our evaluation of the RPN, we generate three versions of the classification pipeline: $CLS_{conv3}$, $CLS_{conv4}$ and $CLS_{conv5}$.
$CLS_{conv5}$ is identical in architecture to the default pipeline described in~\cite{girshick_2015_fastrcnn}.
The other two networks are similar:
They only differ in the feature map that they are based on, and the normalization layer described in chapter~\ref{sec:small_objects_in_faster_r-cnn}.
During training, we only train the fully-connected layers and compare these results to a network where all layers are optimized ($CLS_{conv5} \text{(all)}$).

We train each of these networks on all of the training sets $F_{train,a,b}$ for $40000$ iterations, an initial learning rate of $\mu=0.01$ with a reduction by a factor $\gamma=0.1$ after $30000$ iterations. 
Each of those models in evaluated according to their mean average precision (mAP) on all the test sets $F_{test,x}$ where $a \leq x \leq b$.
Since the ranges of object sizes between the training sets overlap with each other, we obtain multiple mAP values for each object size $x$ -- represented by the test set $F_{test,x}$.
We take the maximum mAP for each version of the classification pipeline.
Since we are solely interested in the performance of the classification stage, we need to eliminate any potential influences between RPN and classification.
We therefore assume a perfect RPN for this experiment and evaluate our networks using groundtruth bounding boxes as object proposals.

\begin{figure}
\centering
\includegraphics[scale=1.0]{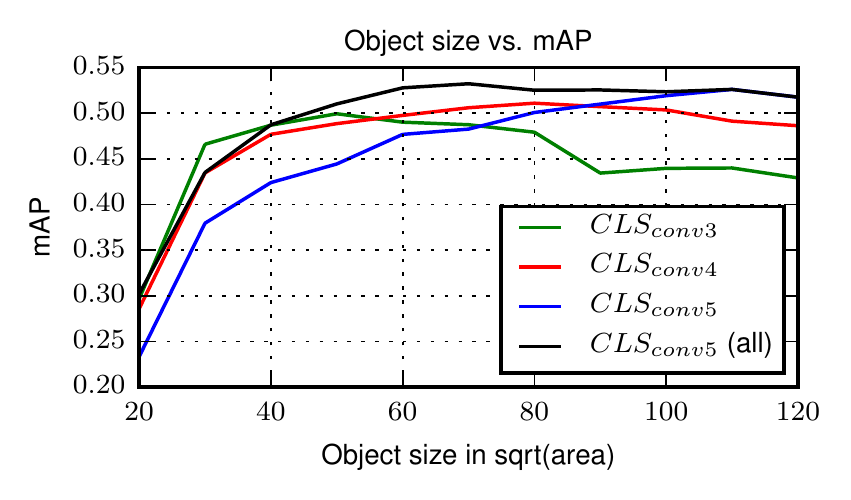}
\caption{Performance of the classification pipeline by size. The performance for \emph{conv5} features drops noticeably for small object sizes. However, a full optimization (\emph{conv5} (all)) is able to adapt to a wide range of scales.}
\label{fig:results_cls_performance_by_size}
\end{figure}

Figure~\ref{fig:results_cls_performance_by_size} shows the results of this experiment.
Unsurprisingly, the classification performance generally declines for small object instances across all feature maps.
When only the fully connected layers are optimized we see a similar trend as in the region proposal network.
For example, the $CLS_{conv3}$ network delivers a better performance for small objects than the $CLS_{conv5}$ network.
When we allow all layers to adapt, this effect disappears.
This behavior shows that DCNNs can potentially adapt to a wide range of input scales.
However, it also suggests that once the network is trained, the filters in each feature map are optimized for a particular scale.

On the other hand, it is generally accepted that deeper layers are more specific in their activations to certain stimuli and are therefore better suited for computer vision tasks.
Ideally, we want to have the best of both worlds: We want to select the feature map at which the filters are best suited for the scale of the object, but we also want high-quality deep features.

\section{An integrated detection pipeline}
\label{sec:an_interated_detection_pipeline}
While our anchor set has the potential to improve object proposals in most detection pipelines which utilize an RPN, we want to improve object proposals even further. 
In our previous experiments we have shown that the resolution of the feature map can have a strong influence on the performance of the RPN, particularly for small anchors.
In the following, we want to explore, whether the performance can be improved further by incorporating information from deeper feature maps.

In the previous chapter we have also examined the scale behavior of the classification stage and have observed a similar behavior.
But the behavior of both stages differs in a key point:
The classification stage is able to accommodate a wide range of object sizes when all layers of the network are allowed to adapt.
This is not the case for the RPN: Even when all layers are allowed to adapt, the scale of an object still determines which feature map is most effective in predicting the ROI.
In other words, the performance of the RPN for small objects is mainly limited by the resolution of the anchor grid which is directly dependent on the resolution of the feature map.

However, the ability of the classification stage to adapt to a wide range of object sizes does not mean that it cannot benefit from information from earlier layers.
In fact, our experiments described in chapter \ref{sec:roi_classification_of_small_objects} suggest that once the network has been trained on objects of a certain size, the object size determines the feature map which is most effective for classification.

\begin{figure}
\centering
\includegraphics[scale=1.0]{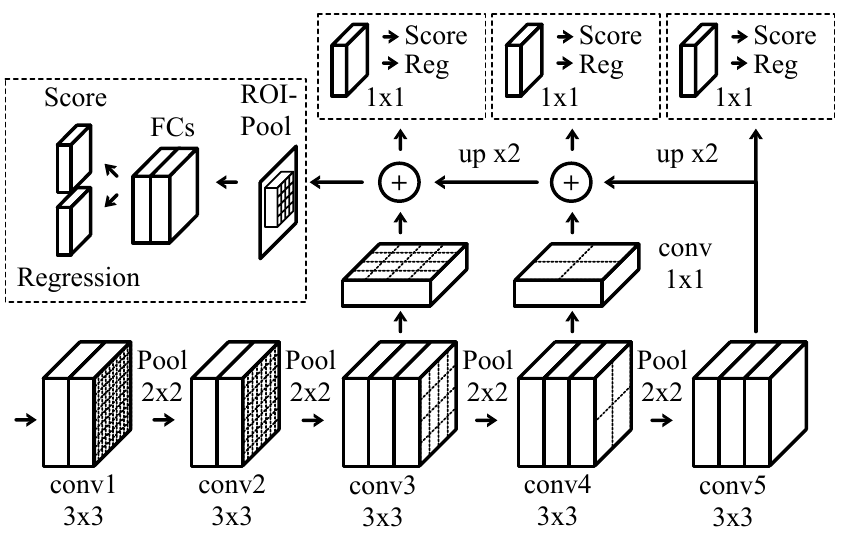}
\caption{Our modified architecture (FPN16) which makes use of feature maps on multiple scales (batch normalization layers are not shown). In addition to the bottom-up feature extraction we add a top-down path which summarizes multi-scale feature representations into a single feature map.}
\label{fig:encoder_decoder_architecture}
\end{figure}

In order to take advantage of the information contained in earlier feature maps we modify our network architecture in the following way (inspired by Feature Pyramid Networks~\cite{lin_2016_fpn}):
In addition to the bottom-up path which extracts the feature hierarchy, we introduce a second top-down path which allows the network to combine feature maps from multiple scales.
This architecture is illustrated in Figure \ref{fig:encoder_decoder_architecture}.
High-level feature maps are upscaled by a factor of $2$ to fit the resolution of the next lower-level feature map using bilinear interpolation.
The lower-level features are allowed to adapt through a $1 \times 1$~convolution before being combined with the upscaled feature map by summation.
This top-down path extends from the \emph{conv5} feature map down to the \emph{conv3} feature map.

We attach RPN modules at every intermediate feature map generated on the top-down path.
Each module is responsible for generating object proposals suitable for its scale.
In accordance with our observations from Figure~\ref{fig:result_rpn_performance_by_anchor_and_size} we use the following assignment of anchor sizes to feature maps:

Anchors with a side length $\leq 45px$ are assigned to the RPN attached to the \emph{conv3} feature map.
All anchors with a side length in the interval $[45px,90px]$ are assigned to the RPN attached to the \emph{conv4} feature map.
Finally, all anchors scales $\geq 90px$ are assigned to the \emph{conv5} feature map.

For generating object proposals we first take the individual proposals from each individual RPN on which we perform an individual non-maximum suppression -- suppressing boxes with an IoU $\geq 0.7$.
The remaining proposals are merged and another non-maximum suppression is applied.

We attach the fully-connected layers of the classification stage at the feature map generated at the end of the top-down path.
In principle, we could also extend the model to train multiple classifiers for separate scales.
However, the vast majority ($> 1*10^8$) of learnable parameters of the VGG16 architecture are concentrated in the fully-connected layers of the network.
A separate classifier for each feature map would inflate the already large model disproportionately.
Since our previous experiments have shown the ability of the classifier to adapt to a wide range of objects, we do not consider this inflation justifiable.

As a sidenote on the implementation: 
Simply attaching additional arms to the existing VGG16 architecture proved unsuccessful.
Since the original network operates on unnormalized input data, the magnitude of the activations varies strongly across the different layers.
We have observed that fine-tuning such a network with added arms can destroy the pre-trained weights.
In order to be able to make these extensions to the VGG16 network we also need to modify its base model.
We introduce batch normalization~\cite{ioffe_2015_batch_normalization} layers after every block of convolutions, that is after the \emph{conv1}, \emph{conv2}, etc.
Additionally, we apply batch normalization directly on the network input, effectively standardizing the input data.
We pre-train the network on the Imagenet~\cite{deng_imagenet_2009} dataset before fine-tuning it on FlickrLogos.

\section{Evaluation}
\label{sec:evaluation}
We evaluate the effectiveness of our approach separately for both stages of the pipeline along two dimensions:
(1) The architecture of the underlying network and (2) the anchor set used to generate object proposals.
We will refer to the unmodified original architecture as \emph{VGG16} which uses only features from the \emph{conv5} feature map. 
\emph{FPN16} refers to our modified architecture which uses information from multiple scales using the top-down path. 
After extracting object proposals on the FlickrLogos dataset ($n=2000$) we hold these proposals fixed and use them to train and evaluate the classification pipeline separately.

We first evaluate the impact of our extended anchor set on the unmodified original pipeline.
For this purpose we evaluate three sets of anchors:
As our baseline we evaluate the default anchor set $\mathcal{A}_{orig}$ which is used by the original Faster R-CNN implementation consisting of the scales $\mathcal{A}_{orig} = \left\lbrace 128, 256, 512 \right\rbrace$.
$\mathcal{A}_{ext} = \left\lbrace 32, 64, 128, 256 \right\rbrace$ refers to the default anchor set which has been adapted for the size distribution of the FlickrLogos dataset (Figure~\ref{fig:flickr_sizedist_orig}) but is a straightforward extension using the default powers-of-two scheme.
Finally, we evaluate an anchor set $\mathcal{A}_{prop}$ which implements our theoretical considerations from chapter~\ref{sec:region_proposals_with_small_objects} which aims to provide a more complete coverage of object scales. $\mathcal{A}_{prop} = \left\lbrace 32, 45, 64, 90, 128, 256 \right\rbrace$.

\begin{table}
\centering
\begin{tabularx}{\columnwidth}{XXXXX}
\hline
Architecture & Anchor set & RPN & CLS & max. Recall \\
\hline
\hline
VGG16 & $\mathcal{A}_{orig}$         & 0.52 & 0.51 & 0.56 \\
\hline
VGG16 & $\mathcal{A}_{ext}$          & 0.66 & 0.62 & 0.72 \\
VGG16 & $\mathcal{A}_{prop}$         & 0.68 & 0.66 & \textbf{0.76} \\
\hline
FPN16 & $\mathcal{A}_{ext}$   & 0.69 & 0.66 & 0.75 \\
FPN16 & $\mathcal{A}_{prop}$  & \textbf{0.71} & \textbf{0.67} & \textbf{0.76} \\
\hline
\end{tabularx}
\vspace{0.2cm}
\caption{Evaluation of RPN and classification performance of both the original architecture and the proposed multiscale architecture for different anchor sets. RPN performance is measured in MABO and classification performance (CLS) is measured in mAP.}
\label{tab:results_rpn_cls}
\end{table}

The results of this evaluation is shown in Table~\ref{tab:results_rpn_cls}.
Surprisingly, the original anchor set performs quite well, despite the fact that it has not been tuned to the size distribution of the FlickrLogos dataset.
However, the strong improvement of the extended anchor set makes it is quite clear that the orignal RPN is missing many small objects.
Our proposed anchor set is able to achieve an even better performance which indicates that the powers-of-two scheme is indeed unable to find an anchor with sufficient overlap for some objects.

Table~\ref{tab:results_rpn_cls} also shows the benefit of of multi-resolution feature maps.
We evaluate \emph{FPN16} on the $\mathcal{A}_{ext}$ and the $\mathcal{A}_{prop}$ anchor sets.
In both cases we are able to achieve a substantial improvement in the performance of the RPN.
Unsurprisingly, the improved object proposals also have a positive impact on the overall detection performance.
However, the relationship between MABO and mAP is not linear.
This is probably due to the fact that small objects typically are detected with a lower confidence score than large objects.

\begin{figure}
\centering
\includegraphics[scale=1.0]{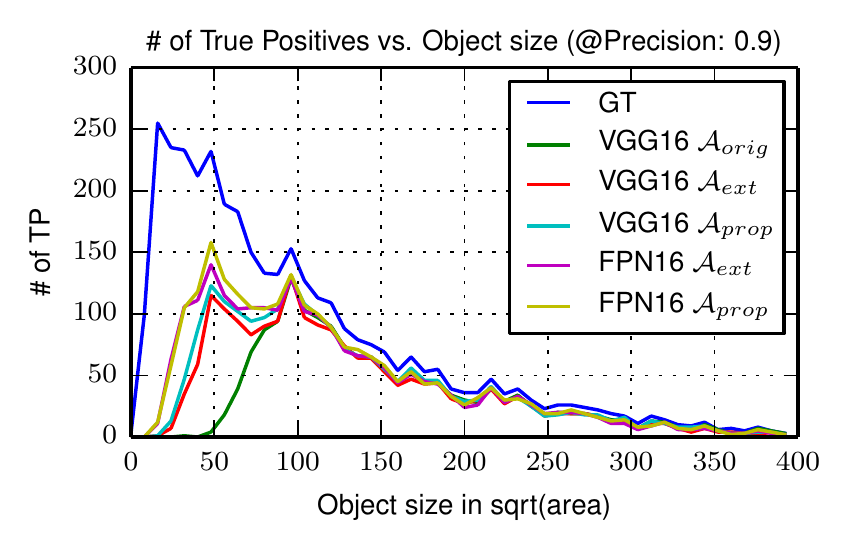}
\caption{True positives as a function of groundtruth instance size on the FlickrLogos test set. An improved detection of small objects is the source of the overall performance increase as measured by the mAP.}
\label{fig:tp_by_size}
\end{figure}

\begin{figure*}
\centering
\includegraphics[scale=0.365]{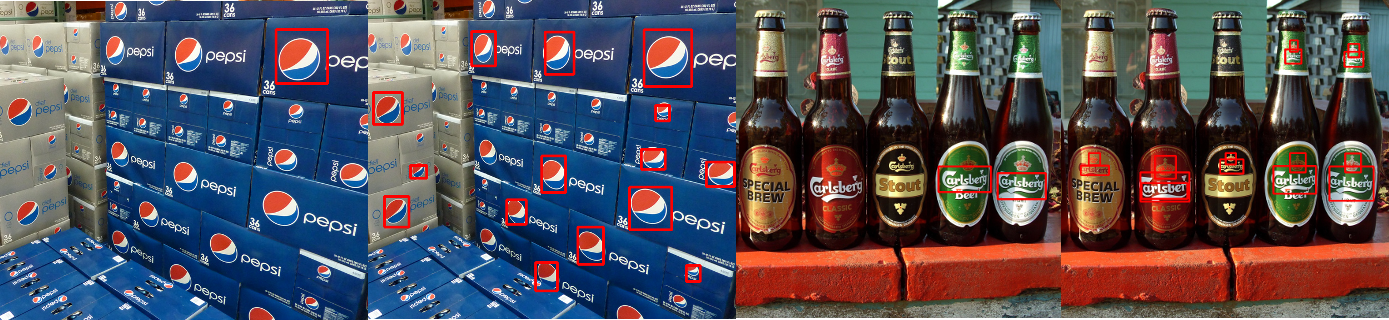}
\caption{(left) Example detections from the FlickrLogos dataset, comparing detections using the $\mathcal{A}_{orig}$ anchor set (VGG16). (right) Example detection using the $\mathcal{A}_{prop}$ anchor set (FPN16). The detection of small logo instances can be visibly improved using our proposed anchor set combined with classification using multiple feature maps.}
\label{fig:example_detections1}
\end{figure*}

Figure~\ref{fig:example_detections1} shows two images with example detections while Figure~\ref{fig:tp_by_size} illustrated the distribution of true positives across all object sizes on the FlickrLogos test set for the approaches evaluated in Table~\ref{tab:results_rpn_cls}.
It is clear, that the overall performance increase as measured by the mAP mostly originates in an improved detection of small objects.

\section{Conclusion}
We have performed a theoretical analysis of the region proposal stage and derived a relationship between feature map resolution and the minimum object size which can reasonably be detected, assuming a perfect classifier.

Our experiments of the scale behavior of the RPN on different feature maps have shown that feature map resolution plays an important role in reliably detecting small object instances.
Small objects can often be detected more accurately on earlier feature maps despite the fact that these features might not be as expressive than features from deeper layers.
This behavior persists, even when all layers of the network are being optimized.

We performed similar experiments on the classification stage and found it to exhibit the same behavior once the filters are fixed. 
However, we have also observed that the filters in the convolutional layers can adapt to a wide range of scales when given the chance.

Finally we have integrated our findings into the Faster R-CNN framework using an improved network architecture based on Feature Pyramid Networks.
We were able to show that integrating features from multiple feature maps while at the same time selecting a suitable resolution for generating proposals can improve the performance of the RPN considerably.

\section{Acknowledgements}
This work was funded by GfK Verein.
The authors would like to thank Carolin Kaiser, Holger Dietrich and Raimund Wildner for the great collaboration. 
Especially, we would like to express our gratitude for their help in re-annotating the FlickrLogos dataset.

\bibliographystyle{ACM-Reference-Format}
\bibliography{bibliography} 


\begin{thebibliography}{00}


\ifx \showCODEN    \undefined \def \showCODEN     #1{\unskip}     \fi
\ifx \showDOI      \undefined \def \showDOI       #1{#1}\fi
\ifx \showISBNx    \undefined \def \showISBNx     #1{\unskip}     \fi
\ifx \showISBNxiii \undefined \def \showISBNxiii  #1{\unskip}     \fi
\ifx \showISSN     \undefined \def \showISSN      #1{\unskip}     \fi
\ifx \showLCCN     \undefined \def \showLCCN      #1{\unskip}     \fi
\ifx \shownote     \undefined \def \shownote      #1{#1}          \fi
\ifx \showarticletitle \undefined \def \showarticletitle #1{#1}   \fi
\ifx \showURL      \undefined \def \showURL       {\relax}        \fi
\providecommand\bibfield[2]{#2}
\providecommand\bibinfo[2]{#2}
\providecommand\natexlab[1]{#1}
\providecommand\showeprint[2][]{arXiv:#2}

\bibitem[\protect\citeauthoryear{Bell, Zitnick, Bala, and Girshick}{Bell
  et~al\mbox{.}}{2016}]%
        {bell_inside_outside_net_2016}
\bibfield{author}{\bibinfo{person}{S. Bell}, \bibinfo{person}{C.~L. Zitnick},
  \bibinfo{person}{K. Bala}, {and} \bibinfo{person}{R. Girshick}.}
  \bibinfo{year}{2016}\natexlab{}.
\newblock \showarticletitle{Inside-Outside Net: Detecting Objects in Context
  with Skip Pooling and Recurrent Neural Networks}. In \bibinfo{booktitle}{{\em
  2016 IEEE Conference on Computer Vision and Pattern Recognition (CVPR)}}.
  \bibinfo{pages}{2874--2883}.
\newblock
\showDOI{%
\url{https://doi.org/10.1109/CVPR.2016.314}}


\bibitem[\protect\citeauthoryear{Bianco, Buzzelli, Mazzini, and
  Schettini}{Bianco et~al\mbox{.}}{2017}]%
        {bianco_2017_deep_learning_for_logo_recognition}
\bibfield{author}{\bibinfo{person}{S. Bianco}, \bibinfo{person}{M. Buzzelli},
  \bibinfo{person}{D. Mazzini}, {and} \bibinfo{person}{R. Schettini}.}
  \bibinfo{year}{2017}\natexlab{}.
\newblock \showarticletitle{Deep Learning for Logo Recognition}.
\newblock \bibinfo{journal}{{\em CoRR\/}}  \bibinfo{volume}{abs/1701.02620}
  (\bibinfo{year}{2017}).
\newblock
\showURL{%
\url{http://arxiv.org/abs/1701.02620}}


\bibitem[\protect\citeauthoryear{Boser, Guyon, and Vapnik}{Boser
  et~al\mbox{.}}{1992}]%
        {boser_1992_support_vector_machine}
\bibfield{author}{\bibinfo{person}{Bernhard~E. Boser},
  \bibinfo{person}{Isabelle~M. Guyon}, {and} \bibinfo{person}{Vladimir~N.
  Vapnik}.} \bibinfo{year}{1992}\natexlab{}.
\newblock \showarticletitle{A Training Algorithm for Optimal Margin
  Classifiers}. In \bibinfo{booktitle}{{\em Proceedings of the Fifth Annual
  Workshop on Computational Learning Theory}} {\em (\bibinfo{series}{COLT
  '92})}. \bibinfo{publisher}{ACM}, \bibinfo{address}{New York, NY, USA},
  \bibinfo{pages}{144--152}.
\newblock
\showISBNx{0-89791-497-X}
\showDOI{%
\url{https://doi.org/10.1145/130385.130401}}


\bibitem[\protect\citeauthoryear{Deng, Dong, Socher, Li, Li, and Fei-Fei}{Deng
  et~al\mbox{.}}{2009}]%
        {deng_imagenet_2009}
\bibfield{author}{\bibinfo{person}{J. Deng}, \bibinfo{person}{W. Dong},
  \bibinfo{person}{R. Socher}, \bibinfo{person}{L.~J. Li}, \bibinfo{person}{Kai
  Li}, {and} \bibinfo{person}{Li Fei-Fei}.} \bibinfo{year}{2009}\natexlab{}.
\newblock \showarticletitle{ImageNet: A large-scale hierarchical image
  database}. In \bibinfo{booktitle}{{\em 2009 IEEE Conference on Computer
  Vision and Pattern Recognition}}. \bibinfo{pages}{248--255}.
\newblock
\showISSN{1063-6919}
\showDOI{%
\url{https://doi.org/10.1109/CVPR.2009.5206848}}


\bibitem[\protect\citeauthoryear{Eggert, Winschel, Zecha, and Lienhart}{Eggert
  et~al\mbox{.}}{2016}]%
        {eggert_saliency_guided_magnification_2016}
\bibfield{author}{\bibinfo{person}{C. Eggert}, \bibinfo{person}{A. Winschel},
  \bibinfo{person}{D. Zecha}, {and} \bibinfo{person}{R. Lienhart}.}
  \bibinfo{year}{2016}\natexlab{}.
\newblock \showarticletitle{Saliency-guided Selective Magnification for Company
  Logo Detection}. In \bibinfo{booktitle}{{\em 2016 International Conference on
  Pattern Recognition (ICPR)}}.
\newblock


\bibitem[\protect\citeauthoryear{Everingham, Gool, Williams, Winn, and
  Zisserman}{Everingham et~al\mbox{.}}{2010}]%
        {everingham_2010_voc}
\bibfield{author}{\bibinfo{person}{M. Everingham}, \bibinfo{person}{L.~Van
  Gool}, \bibinfo{person}{C.~K.~I. Williams}, \bibinfo{person}{J. Winn}, {and}
  \bibinfo{person}{A. Zisserman}.} \bibinfo{year}{2010}\natexlab{}.
\newblock \showarticletitle{The Pascal Visual Object Classes (VOC) Challenge}.
\newblock \bibinfo{journal}{{\em International Journal of Computer Vision\/}}
  \bibinfo{volume}{88}, \bibinfo{number}{2} (\bibinfo{date}{June}
  \bibinfo{year}{2010}), \bibinfo{pages}{303--338}.
\newblock


\bibitem[\protect\citeauthoryear{Girshick}{Girshick}{2015}]%
        {girshick_2015_fastrcnn}
\bibfield{author}{\bibinfo{person}{R. Girshick}.}
  \bibinfo{year}{2015}\natexlab{}.
\newblock \showarticletitle{Fast R-CNN}. In \bibinfo{booktitle}{{\em IEEE
  International Conference on Computer Vision}}. \bibinfo{pages}{1440--1448}.
\newblock
\showDOI{%
\url{https://doi.org/10.1109/ICCV.2015.169}}


\bibitem[\protect\citeauthoryear{Girshick, Donahue, Darrell, and
  Malik}{Girshick et~al\mbox{.}}{2014}]%
        {girshick_2014_rcnn}
\bibfield{author}{\bibinfo{person}{R. Girshick}, \bibinfo{person}{J. Donahue},
  \bibinfo{person}{T. Darrell}, {and} \bibinfo{person}{J. Malik}.}
  \bibinfo{year}{2014}\natexlab{}.
\newblock \showarticletitle{Rich Feature Hierarchies for Accurate Object
  Detection and Semantic Segmentation}. In \bibinfo{booktitle}{{\em IEEE
  Conference on Computer Vision and Pattern Recognition}}.
  \bibinfo{pages}{580--587}.
\newblock
\showDOI{%
\url{https://doi.org/10.1109/CVPR.2014.81}}


\bibitem[\protect\citeauthoryear{Hariharan, Arbelaez, Girshick, and
  Malik}{Hariharan et~al\mbox{.}}{2016}]%
        {hariharan_2016_hypercolumns}
\bibfield{author}{\bibinfo{person}{B. Hariharan}, \bibinfo{person}{P.
  Arbelaez}, \bibinfo{person}{R. Girshick}, {and} \bibinfo{person}{J. Malik}.}
  \bibinfo{year}{2016}\natexlab{}.
\newblock \showarticletitle{Object Instance Segmentation and Fine-Grained
  Localization using Hypercolumns}.
\newblock \bibinfo{journal}{{\em IEEE Transactions on Pattern Analysis and
  Machine Intelligence\/}} \bibinfo{volume}{PP}, \bibinfo{number}{99}
  (\bibinfo{year}{2016}).
\newblock
\showISSN{0162-8828}
\showDOI{%
\url{https://doi.org/10.1109/TPAMI.2016.2578328}}


\bibitem[\protect\citeauthoryear{Ioffe and Szegedy}{Ioffe and Szegedy}{2015}]%
        {ioffe_2015_batch_normalization}
\bibfield{author}{\bibinfo{person}{S. Ioffe} {and} \bibinfo{person}{C.
  Szegedy}.} \bibinfo{year}{2015}\natexlab{}.
\newblock \showarticletitle{Batch Normalization: Accelerating Deep Network
  Training by Reducing Internal Covariate Shift}. In \bibinfo{booktitle}{{\em
  International Conference on Machine Learning}}. \bibinfo{pages}{448--465}.
\newblock


\bibitem[\protect\citeauthoryear{Lin, Doll\'{a}r, Girshick, He, Hariharan, and
  Belongie}{Lin et~al\mbox{.}}{2016}]%
        {lin_2016_fpn}
\bibfield{author}{\bibinfo{person}{Tsung-Yi Lin}, \bibinfo{person}{Piotr
  Doll\'{a}r}, \bibinfo{person}{Ross Girshick}, \bibinfo{person}{Kaiming He},
  \bibinfo{person}{Bharath Hariharan}, {and} \bibinfo{person}{Serge Belongie}.}
  \bibinfo{year}{2016}\natexlab{}.
\newblock \showarticletitle{Feature Pyramid Networks for Object Detection}.
\newblock \bibinfo{journal}{{\em arXiv preprint arXiv:1612.03144\/}}
  (\bibinfo{year}{2016}).
\newblock


\bibitem[\protect\citeauthoryear{Liu, Anguelov, Erhan, Szegedy, Reed, Fu, and
  Berg}{Liu et~al\mbox{.}}{2016}]%
        {wei_2016_ssd}
\bibfield{author}{\bibinfo{person}{Wei Liu}, \bibinfo{person}{Dragomir
  Anguelov}, \bibinfo{person}{Dumitru Erhan}, \bibinfo{person}{Christian
  Szegedy}, \bibinfo{person}{Scott Reed}, \bibinfo{person}{Cheng-Yang Fu},
  {and} \bibinfo{person}{Alexander~C. Berg}.} \bibinfo{year}{2016}\natexlab{}.
\newblock \showarticletitle{SSD: Single Shot MultiBox Detector}.
\newblock  (\bibinfo{year}{2016}).
\newblock
\showURL{%
\url{http://arxiv.org/abs/1512.02325}}


\bibitem[\protect\citeauthoryear{Oliveira, Frazao, Pimentel, and
  Ribeiro}{Oliveira et~al\mbox{.}}{2016}]%
        {oliveira_2016_automatic_graphic_logo_detection}
\bibfield{author}{\bibinfo{person}{G. Oliveira}, \bibinfo{person}{X. Frazao},
  \bibinfo{person}{A. Pimentel}, {and} \bibinfo{person}{B. Ribeiro}.}
  \bibinfo{year}{2016}\natexlab{}.
\newblock \showarticletitle{Automatic graphic logo detection via Fast
  Region-based Convolutional Networks}. In \bibinfo{booktitle}{{\em 2016
  International Joint Conference on Neural Networks (IJCNN)}}.
  \bibinfo{pages}{985--991}.
\newblock
\showDOI{%
\url{https://doi.org/10.1109/IJCNN.2016.7727305}}


\bibitem[\protect\citeauthoryear{Redmon, Divvala, Girshick, and Farhadi}{Redmon
  et~al\mbox{.}}{2016}]%
        {redmon_2016_yolo}
\bibfield{author}{\bibinfo{person}{J. Redmon}, \bibinfo{person}{S. Divvala},
  \bibinfo{person}{R. Girshick}, {and} \bibinfo{person}{A. Farhadi}.}
  \bibinfo{year}{2016}\natexlab{}.
\newblock \showarticletitle{You Only Look Once: Unified, Real-Time Object
  Detection}. In \bibinfo{booktitle}{{\em 2016 IEEE Conference on Computer
  Vision and Pattern Recognition (CVPR)}}. \bibinfo{pages}{779--788}.
\newblock
\showDOI{%
\url{https://doi.org/10.1109/CVPR.2016.91}}


\bibitem[\protect\citeauthoryear{Ren, He, Girshick, and Sun}{Ren
  et~al\mbox{.}}{2016}]%
        {ren_2015_faster_rcnn}
\bibfield{author}{\bibinfo{person}{S. Ren}, \bibinfo{person}{K. He},
  \bibinfo{person}{R. Girshick}, {and} \bibinfo{person}{J. Sun}.}
  \bibinfo{year}{2016}\natexlab{}.
\newblock \showarticletitle{Faster R-CNN: Towards Real-Time Object Detection
  with Region Proposal Networks}.
\newblock \bibinfo{journal}{{\em IEEE Transactions on Pattern Analysis and
  Machine Intelligence\/}} (\bibinfo{year}{2016}).
\newblock
\showISSN{0162-8828}
\showDOI{%
\url{https://doi.org/10.1109/TPAMI.2016.2577031}}


\bibitem[\protect\citeauthoryear{Romberg, Pueyo, R., and van Zwol}{Romberg
  et~al\mbox{.}}{2011}]%
        {Romberg_2011}
\bibfield{author}{\bibinfo{person}{S. Romberg}, \bibinfo{person}{L.~G. Pueyo},
  \bibinfo{person}{Lienhart R.}, {and} \bibinfo{person}{R. van Zwol}.}
  \bibinfo{year}{2011}\natexlab{}.
\newblock \showarticletitle{Scalable logo recognition in real-world images}. In
  \bibinfo{booktitle}{{\em ACM International Conference on Multimedia
  Retrieval}} {\em (\bibinfo{series}{ICMR '11})}. \bibinfo{publisher}{ACM},
  Article \bibinfo{articleno}{25}, \bibinfo{numpages}{8}~pages.
\newblock
\showISBNx{978-1-4503-0336-1}
\showDOI{%
\url{https://doi.org/10.1145/1991996.1992021}}


\bibitem[\protect\citeauthoryear{Shelhamer, Long, and Darrell}{Shelhamer
  et~al\mbox{.}}{2016}]%
        {shelhamer_2016_semantic_image_segmentation}
\bibfield{author}{\bibinfo{person}{E. Shelhamer}, \bibinfo{person}{J. Long},
  {and} \bibinfo{person}{T. Darrell}.} \bibinfo{year}{2016}\natexlab{}.
\newblock \showarticletitle{Fully Convolutional Networks for Semantic
  Segmentation}.
\newblock \bibinfo{journal}{{\em IEEE Transactions on Pattern Analysis and
  Machine Intelligence\/}} \bibinfo{volume}{PP}, \bibinfo{number}{99}
  (\bibinfo{year}{2016}).
\newblock
\showISSN{0162-8828}
\showDOI{%
\url{https://doi.org/10.1109/TPAMI.2016.2572683}}


\bibitem[\protect\citeauthoryear{Simonyan and Zisserman}{Simonyan and
  Zisserman}{2015}]%
        {simonyan_2015_very_deep_networks}
\bibfield{author}{\bibinfo{person}{K. Simonyan} {and} \bibinfo{person}{A.
  Zisserman}.} \bibinfo{year}{2015}\natexlab{}.
\newblock \showarticletitle{Very Deep Convolutional Networks for Large-Scale
  Image Recognition}. In \bibinfo{booktitle}{{\em International Conference on
  Learning Representations}}.
\newblock


\bibitem[\protect\citeauthoryear{Uijlings, Sande, Gevers, and
  Smeulders}{Uijlings et~al\mbox{.}}{2013}]%
        {uijlings_2013_selsearch}
\bibfield{author}{\bibinfo{person}{J.~R.~R. Uijlings},
  \bibinfo{person}{K.~E.~A. Sande}, \bibinfo{person}{T. Gevers}, {and}
  \bibinfo{person}{A.~W.~M. Smeulders}.} \bibinfo{year}{2013}\natexlab{}.
\newblock \showarticletitle{Selective Search for Object Recognition}.
\newblock \bibinfo{journal}{{\em International Journal of Computer Vision\/}}
  \bibinfo{volume}{104}, \bibinfo{number}{2} (\bibinfo{year}{2013}),
  \bibinfo{pages}{154--171}.
\newblock
\showISSN{1573-1405}
\showDOI{%
\url{https://doi.org/10.1007/s11263-013-0620-5}}


\bibitem[\protect\citeauthoryear{Zitnick and Doll{\'a}r}{Zitnick and
  Doll{\'a}r}{2014}]%
        {zitnick_2014_edge_boxes}
\bibfield{author}{\bibinfo{person}{L. Zitnick} {and} \bibinfo{person}{P.
  Doll{\'a}r}.} \bibinfo{year}{2014}\natexlab{}.
\newblock \showarticletitle{Edge Boxes: Locating Object Proposals from Edges}.
  In \bibinfo{booktitle}{{\em ECCV}}. \bibinfo{publisher}{European Conference
  on Computer Vision}.
\newblock


\end{thebibliography}

\end{document}